\documentclass[journal]{IEEEtran}
\usepackage{amsmath}
\usepackage{amssymb}
\usepackage{mathrsfs}
\usepackage{cite} 
\usepackage{booktabs}
\usepackage[table,xcdraw]{xcolor}
\usepackage[normalem]{ulem}
\useunder{\uline}{\ul}{}
\usepackage{multirow}
\usepackage{graphicx}
\usepackage{subfigure} 
\usepackage{caption}
\usepackage[ruled,linesnumbered]{algorithm2e} 
\usepackage{hyperref}
\usepackage{url}
\usepackage{epstopdf}
\usepackage{lineno}
\begin{document}
	

	\title{MS-Former: Memory-Supported Transformer for Weakly Supervised Change Detection with Patch-Level Annotations}

	\author{Zhenglai Li, 
		Chang Tang, ~\IEEEmembership{Senior Member,~IEEE,}
		Xinwang Liu, ~\IEEEmembership{Senior Member,~IEEE,}
		Changdong Li,
		Xianju Li,\\
		Wei Zhang,
		\thanks{Z. Li,  C. Tang, and X. Li are with the school of computer, China University of Geosciences, Wuhan, China. 
			(E-mail: \{yuezhenguan, tangchang, ddwhlxj\}@cug.edu.cn).}
		\thanks{X. Liu is with the school of computer, National University of Defense Technology, Changsha 410073, China.
			(E-mail:  xinwangliu@nudt.edu.cn).}
		\thanks{C. Li is with the faculty of engineering, China University of Geosciences, Wuhan, China. 
				(E-mail: lichangdong@cug.edu.cn).}
		\thanks{W. Zhang is with Shandong Provincial Key Laboratory of Computer Networks, Shandong Computer Science Center (National Supercomputing Center in Jinan), Qilu University of Technology (Shandong Academy of Sciences), Jinan 250000, China.
			(E-mail:  wzhang@qlu.edu.cn).}
		\thanks{Manuscript received April 19, 2021; revised August 16, 2021. This work was supported in part by the National Science Foundation of China under Grant 62076228, and in part by Natural Science Foundation of Shandong Province under Grant ZR2021LZH001. (Corresponding author: Chang Tang.)}}

	\markboth{Journal of \LaTeX\ Class Files,~Vol.~14, No.~8, August~2015}%
	{Shell \MakeLowercase{\textit{et al.}}: Bare Demo of IEEEtran.cls for IEEE Journals}
	%



	\maketitle
	
	\begin{abstract}
		Fully supervised change detection methods have achieved significant advancements in performance, yet they depend severely on acquiring costly pixel-level labels.  Considering that the patch-level annotations also contain abundant information corresponding to both changed and unchanged objects in bi-temporal images, an intuitive solution is to segment the changes with patch-level annotations.  How to capture the semantic variations associated with the changed and unchanged regions from the patch-level annotations to obtain promising change results is the critical challenge for the weakly supervised change detection task.  In this paper, we propose a memory-supported transformer (MS-Former), a novel framework consisting of a bi-directional attention block (BAB) and a patch-level supervision scheme (PSS) tailored for weakly supervised change detection with patch-level annotations.  More specifically, the BAM captures contexts associated with the changed and unchanged regions from the temporal difference features to construct informative prototypes stored in the memory bank.  On the other hand, the BAM extracts useful information from the prototypes as supplementary contexts to enhance the temporal difference features, thereby better distinguishing changed and unchanged regions.  After that, the PSS guides the network learning valuable knowledge from the patch-level annotations, thus further elevating the performance.  Experimental results on three benchmark datasets demonstrate the effectiveness of our proposed method in the change detection task. The demo code for our work will be publicly available at \url{https://github.com/guanyuezhen/MS-Former}.	
	\end{abstract}
	

	%
	\IEEEpeerreviewmaketitle

	\section{Introduction}
	As a fundamental technique in the area of remote sensing image understanding, change detection (CD) locates the landscape changes and assigns each pixel of co-registered bi-temporal images captured at diverse periods within a given geographical area with binary categories, e.g., changed or unchanged. During the past several years, CD has received increased attention and been applied for a wide range of applications, such as land-use change detecting~\cite{zheng2022changemask, hu2018automatic}, urban development monitoring~\cite{chen2020spatial, pang2023detecting}, global resources monitoring~\cite{yuan2022transformer}, and damage assessment~\cite{li2016landslide, zheng2021building}. 
	
	With the rapid development of deep learning models, numerous approaches tailored specifically for the CD task have been introduced~\cite{shafique2022deep}. However, the impressive performance exhibited by most existing CD methods is heavily depended on the well-annotated annotated pixel-level labels, a process known for its time-consuming and labor-intensive properties. The advent of high-resolution satellites globally has spurred interest in weakly supervised CD techniques, as they offer more cost-effective alternatives, such as image-level labels, to strike a balance between performance and annotation expenditure. Several methods leveraging diverse techniques have been proposed to enhance CD performance guided by image-level annotations. For instance, Andermatt et al.~\cite{andermatt2020weakly} exploited the Conditional Random Field (CRF) to refine the change mask in the hidden layer. Kalita et al.~\cite{kalita2021land} combined the Principal Component Analysis (PCA) and $k$-means clustering algorithm to produce the change maps. Wu et al.~\cite{wu2023fully} introduced a Generative Adversarial Network (GAN) to solve the unsupervised, weakly-supervised, and region-supervision CD in a unified framework. Huang et al.~\cite{huang2023background} proposed a Background-Mixed sample augmentation approach (BGMix) to augment samples with the help of a set of background-changing images. 
	
	\begin{figure*}[t]
		\centering
		\includegraphics[width=0.95\textwidth]{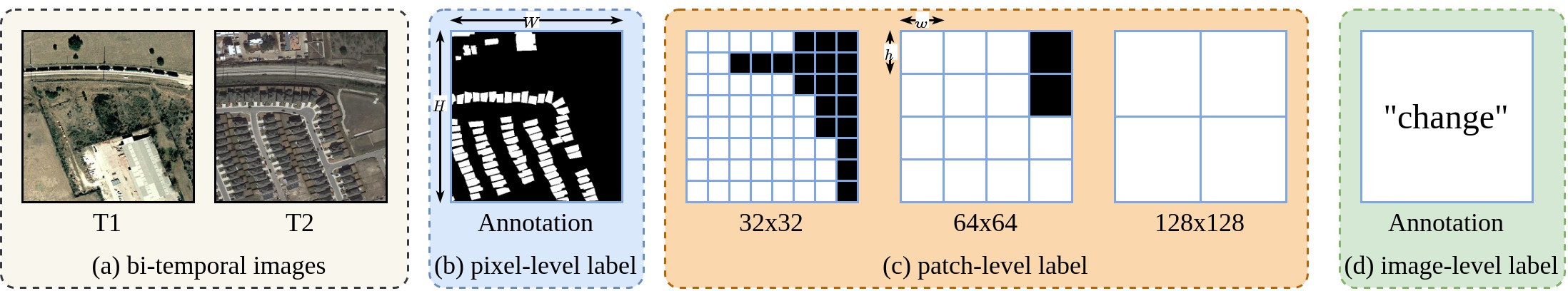}
		\caption{
			Comparison of the pixel-level, image-level, and our patch-level labels for remote sensing change detection.
		}
		\label{fig:patch_level_label}
	\end{figure*}
	
	\begin{figure}[t]
		\centering
		\includegraphics[width=0.48\textwidth]{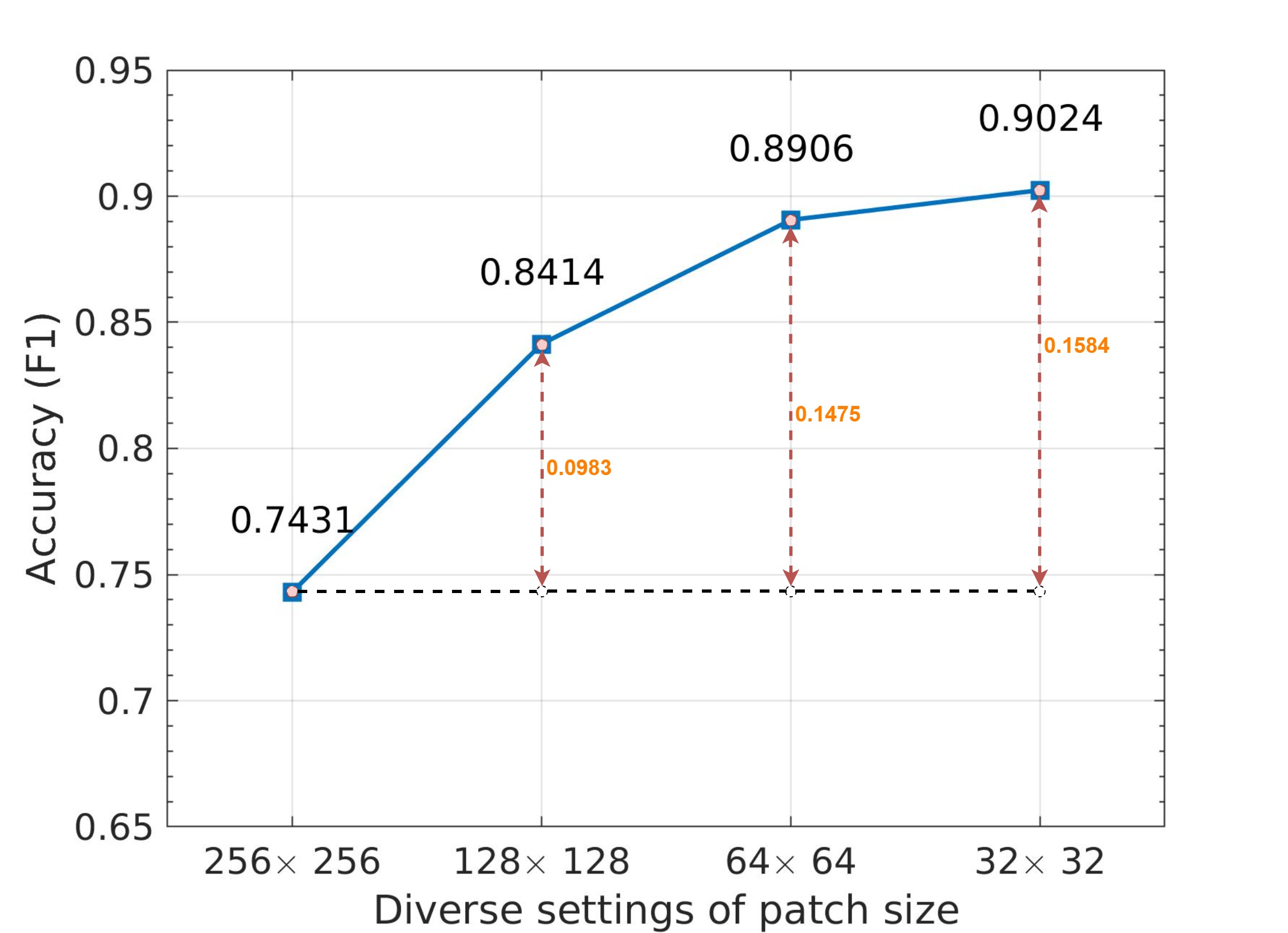}
		\caption{
			Comparison of the change detection performance measured by F1 of our proposed MS-Former using patch-level labels across different patch size settings on the BCDD dataset.
		}
		\label{fig:performance}
	\end{figure}
	
	Fig~\ref{fig:patch_level_label} gives co-registered bi-temporal images with pixel-level, patch-level, and image-level labels. The initial $H\times W$ bi-temporal images are cropped into small patches of size $h\times w$. These patches are assigned binary labels indicating the presence or absence of changes, thereby formulating the patch-level annotations. Notably, as the patch size increases, the patch-level labels align more closely with image-level annotations, while decreasing patch size results in labels close to pixel-wise annotations. In this work, we observe that a slight reduction in patch size substantially enhances change detection performance, as given in Fig~\ref{fig:performance}. Specifically, on the BCDD dataset~\footnote{http://study.rsgis.whu.edu.cn/pages/download/building\_dataset.html}, the F1 scores for change detection under patch sizes $256\times 256$, $128\times 128$, $64\times 64$, and $32\times 32$ are 0.7431, 0.8414, 0.8906, and 0.9014, respectively. Moreover, the F1 scores demonstrate a significant increase of approximately 9.83, 14.75, and 15.84 percentages when transitioning from patch sizes $256\times 256$ to $128\times 128$, $64\times 64$, and $32\times 32$, respectively. This observation suggests the potential of exploring patch-level annotations for remote sensing change detection.
	
	In this paper, we propose a novel memory-supported transformer (MS-Former) for the purpose of weakly supervised CD. Inspired by the recent advantage of the memory bank mechanism in deep learning, we introduce a memory bank to store the prototypes depicting semantic variations associated with the changed and unchanged regions. To this end, our MS-Former incorporates a bi-directional attention block (BAB) designed to enhance context extraction from bi-temporal images in a dual-directional manner. First, the BAB selects salient feature representations within each patch and fuses them with the prototypes stored in the memory bank, progressively capturing the distinctive characteristics of changes across the entire dataset in an online procedure. Second, the BAB extracts supplementary contextual information from the prototypes to refine the temporal difference features, thus facilitating a more effective distinguishing between changed and unchanged regions within bi-temporal images. Concurrently, our patch-level supervision scheme (PSS) delivers patch-level labels at both high and low resolutions, guiding the network in acquiring valuable insights from patch-level annotations and thereby further enhancing its performance.
	
	The main contributions of this work are summarized as follows:
	\begin{enumerate}
		\item We establish a novel benchmark for remote sensing CD incorporating patch-level supervision. The introduced patch-level annotations strike a balance between change detection performance and the associated label annotation costs compared with both image-level and pixel-level labels.
		\item We design a novel memory-supported transformer (MS-Former) tailored for remote sensing change detection under patch-level supervision. The MS-Former integrates a bi-directional attention block (BAB) and patch-level supervision scheme (PSS) to achieve considerable change detection performance.
		\item A comprehensive set of experiments is conducted on three benchmark datasets to assess the effectiveness and superiority of our proposed MS-Former. Considerable experimental results suggest that patch-level supervision could offer an effective solution to the remote sensing CD task.
	\end{enumerate}
	
	The structure of this paper is organized as follows: Section~\ref{sec:sec2} offers a review of related literature concerning remote sensing CD and memory bank mechanisms. In Section~\ref{sec:sec3}, the proposed method is elaborated in detail. Extensive experiments, conducted on three benchmark datasets to assess the performance of the method, along with comprehensive discussions and model analyses, are presented in Section~\ref{sec:sec4}. The paper concludes in Section~\ref{sec:sec5}.
	
	\section{Related Work}\label{sec:sec2}
	\subsection{Change Detection}
	\subsubsection{Fully supervised change detection}
	With the rapid advancement of deep learning techniques, CD methods have made significant strides, particularly under the guidance of pixel-wise supervision
	Daudt et al.~\cite{daudt2018fully}, proposed three end-to-end CD networks based on U-net ~\cite{Unet_2015u}, and explored the effectiveness of three diverse architectures, e.g., FC-EF, FC-Siam-conc, and FC-Siam-diff. Building upon this foundation, subsequent efforts have focused on enhancing CD performance. For instance, the integration of long short-term memory (LSTM) and skip connection is employed to extract more distinguishable multi-level features~\cite{papadomanolaki2021deep}. A full-scale feature fusion approach has been introduced to aggregate information across various feature scales. Additionally, a hybrid attention mechanism is utilized to capture long-range contextual information among bi-temporal features, thereby enhancing the accuracy of CD~\cite{li2022densely}. Li et al.~\cite{Li2022cd} proposed a guided refinement model, which aggregates multi-level features and iteratively refines them, effectively filtering out irrelevant noise from the features. Further insights into recent developments in this field can be found in the comprehensive survey~\cite{shafique2022deep}.
	
	Previous CD methods primarily explore multi-level feature fusion~\cite{zhang2023asymmetric, feng2023change, li2023lightweight}, temporal difference extraction~\cite{zheng2022changemask, lei2021difference, fang2023changer}, and attention mechanisms~\cite{hu2018squeeze, woo2018cbam, wang2018non} to capture distinctive feature representations, thus optimizing performance under the guidance of pixel-level annotations. However, the labor-intensive nature of pixel-wise labeling renders it impractical and inefficient for addressing large-scale remote sensing data challenges.
	
	\subsubsection{Weakly supervised change detection}
	Weakly supervised CD adopt more economical weak supervisions, such as image-level labels, as opposed to intricate pixel-wise annotations, aiming to strike a balance between performance and expenses. Several techniques have been proposed to enhance CD performance relying on image-level annotations. For example, Andermatt et al.\cite{andermatt2020weakly} utilized the Conditional Random Field (CRF) in the hidden layer to refine the change mask. Kalita et al.\cite{kalita2021land} combined Principal Component Analysis (PCA) with the $k$-means clustering algorithm to generate change maps. Wu et al.\cite{wu2023fully} introduced Generative Adversarial Networks (GAN) to tackle unsupervised, weakly-supervised, and region-supervised CD challenges. Huang et al.\cite{huang2023background} devised a sample augmentation approach, integrating background images effectively into the samples. These methods demonstrate the diverse approaches employed to optimize CD performance with the support of image-level annotations, avoiding costly pixel-wise annotations. In contrast to prior approaches utilizing image-level labels, our study introduces novel patch-level annotations with flexible patch size settings, enabling a fine balance between performance and costs.
	
	\subsection{Memory Bank Mechanisms}
	The memory bank mechanism is a widely used technique in deep learning. For instance, Li et al,~\cite{PAU} introduced diverse memory prototypes for each modality to capture its semantic structure for enhancing cross-modal retrieval. Cui et al,~\cite{cai2021appearance} proposed an appearance-motion memory consistency network to make full use of the appearance and motion knowledge in the high-level feature space to perform video anomaly detection. In~\cite{xie2021efficient}, a regional memory network is presented to capture the object cues from the past frames in a local-to-local matching manner to boost the performance of video object segmentation. In recent years, the memory bank mechanism has received much attention in weakly-supervised semantic segmentation~\cite{fan2022memory, zhou2022regional}. Fan et al,~\cite{fan2022memory} utilized the memory bank mechanism to explore cross-image contexts to refine the pseudo-masks. Zhou et al,~\cite{zhou2022regional} introduced a regional memory bank to store the diverse object patterns, which are constrained by a regional semantic contrast regularization to recover the dataset-level semantic structure, leading to improved segmentation performance. In~\cite{ru2022weakly}, a visual words learning strategy is proposed to force the network focus on partial discriminative object regions to generate more accurate Class Activation Maps (CAMs). Inspired by the great progress of the memory bank mechanism in deep learning, we apply the memory to explore the semantic variations associated with the changed and unchanged regions to weakly supervised CD, which is a novel design in this field.
	
	\section{Proposed Method}\label{sec:sec3}
	In this section, we first introduce the weakly supervised CD problem and provide an overview of our proposed MS-Former. Subsequently, we explicate the individual modules of our proposed MS-Former in detail. Finally, we illustrate the training loss of the method.
	
	\subsection{Overview}
	Given a registered pair of images $\mathbf{I}^t \in \mathbb{R}^{H\times W \times 3}, t\in\{1, 2\}$, where $H$, $W$, and $t$ respectively denote the height, width, and temporal order of images, the bi-temporal images can be cropped into multiple none-overlapped paired patches as,
	\begin{equation}
		\mathcal{P}(\mathbf{P}_k^t, k\in\{1, 2, \cdots, K\}) = \mathsf{Crop}(\mathbf{I}^t, t\in\{1, 2\}),
	\end{equation}
	where $\mathbf{P}_k^t \in \mathbb{R}^{h\times w \times 3}, t\in\{1, 2\}$ is the pair of bi-temporal image patches. And, $h$ and $w$ represent the height and width of patch size. Then, a binary indicator is utilized to label whether there are changes in each patch. Such process can be denoted as,
	\begin{equation}\label{eq:patch_level_annotations}
		\mathbf{Y}_k = \begin{cases} 
			\mathbf{1},  & \mbox{if changes exist in }\mathbf{P}_k^1\mbox{ and }\mathbf{P}_k^2, \\
			\mathbf{0}, & \mbox{otherwise}
		\end{cases}
	\end{equation}
	where $\mathbf{Y}_k$ is the $k$-th patch in the patch-level label $\mathbf{Y} \in \mathbb{R}^{H\times W}$. 
	
	As highlighted in the previous section, it is evident that when the patch size is set to match the image size, the patch-level labels align with the image-level labels. As the patch size decreases, the obtained patch-level annotations closely approximate the pixel-level labels. Therefore, fine-tuning the patch size within appropriate ranges enables a balance between change detection performance and the costs associated with label annotations. During the training phase, our model is trained to utilize bi-temporal images with patch-level annotations, enabling precise identification of changes within the bi-temporal images. Notably, in comparison to pixel-wise annotations, the utilization of patch-level annotations significantly reduces the overall labeling costs.
	
	\begin{figure*}[t]
		\centering
		\includegraphics[width=0.9\textwidth]{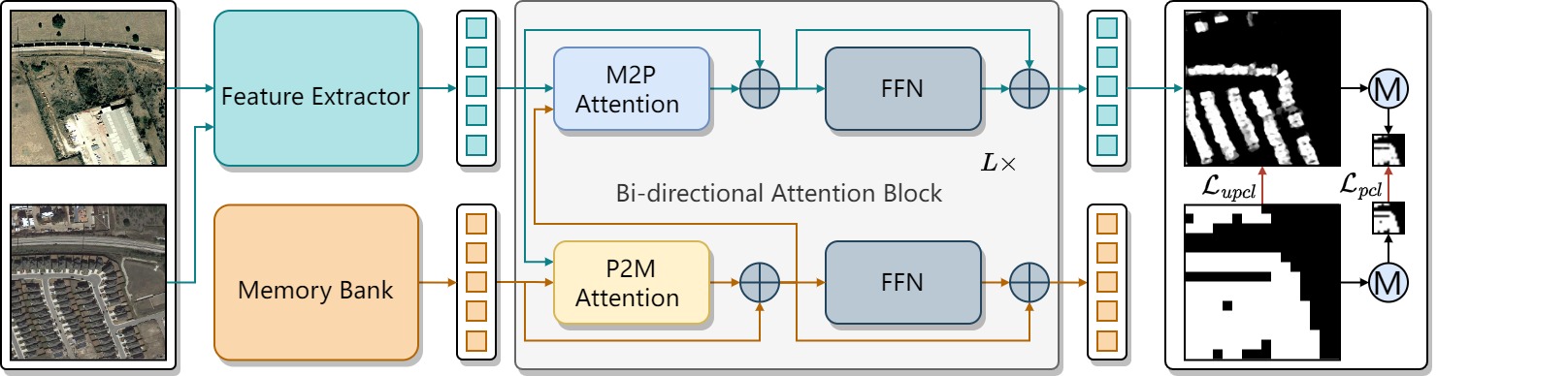}
		\caption{A framework of the proposed MS-Former. 
			Initially, the bi-temporal images pass through a feature extractor to capture the temporal difference features. After that, the temporal difference features and prototypes stored in the memory bank are jointly learned by a series of bi-directional attention blocks. Finally, a patch-level supervision scheme is introduced to guide the network learning from the patch-level annotations.}
		\label{fig:WCDNet}
	\end{figure*}
	
	As illustrated in Fig.~\ref{fig:WCDNet}, the proposed MS-Former consists of three key modules, including a feature extractor, a bi-directional attention block, and a patch-level supervision scheme. The feature extractor is applied to extract temporal difference feature representations from the bi-temporal images. The bi-directional attention block is designed to learn the changes-aware prototypes from the whole training dataset and leverage the changes-aware prototypes to refine the temporal difference features to generate more accurate change maps. Finally, the patch-level supervision scheme is introduced to guide the model learning from the patch-level annotations. More details of the feature extractor, bi-directional attention block, and patch-level supervision scheme will be presented in Subsection \ref{sub:fe}, \ref{sub:bam}, and \ref{sub:pss}, respectively.
	
	\subsection{Feature Extractor}\label{sub:fe}
	The feature extractor contains a weight-shared backbone to obtain multi-level bi-temporal features, denoted as $\mathbf{F}_i^t \in \mathbb{R}^{H/2^{i+1} \times W/2^{i+1} \times c_i}, i\in\{2, 3, 4, 5\}, t\in\{1, 2\}$. Then, we utilize an element-wise subtraction operation to extract temporal difference information, i.e,  $\mathbf{D}_i \in \mathbb{R}^{H/2^{i+1} \times W/2^{i+1} \times c_i}, i\in\{2, 3, 4, 5\}$ from multi-level bi-temporal features. Finally, a decoder is exploited to aggregate the multi-level temporal difference information to obtain a unified feature representation for change detection.
	
	\subsection{Bi-directional Attention Block}\label{sub:bam}
	The memory bank mechanism is a widely used tool in deep learning for exploring the semantic structure of the whole dataset. Inspired by this, we formulate a memory bank to store the semantic variations associated with the changed and unchanged regions. Afterward, a bi-directional attention block (BAB) is introduced to construct the connection between the prototypes stored in the memory bank and the pixel-wise temporal difference feature representations. The goal of BAB is to employ the prototypes, which are learned from the whole dataset, to enhance the temporal feature representations, enabling generate better change maps, as shown in Fig.~\ref{fig:WCDNet}. The BAB is composed of a pixel-to-memory (P2M) attention layer, a memory-to-pixel (M2P) attention layer, and two feed-forward layers~\cite{vaswani2017attention}. In the following content, the P2M and M2P layers will be presented in detail.
	
	\subsubsection{Pixel-to-Memory Attention}
	The memory bank stores the contextual prototypes of changed and unchanged regions. In this part, we introduce pixel-to-memory (P2M) attention, which extracts contexts from the pixel-wise temporal difference feature representations to update the prototypes. Let $\mathbf{P}_s \in \mathbb{R}^{N_p\times C}$ and $\mathbf{M}_s \in \mathbb{R}^{N_m\times C}$ be the inputs of the $s$-th BAM, where $N_p$, $N_m$, and $C$ are the number of temporal difference feature tokens, memory length, and channels, respectively. We first leverage a representative feature extractor to obtain contexts from each patch in the paired bi-temporal images. The adaptive max pooling operation is utilized to extract the most representative feature $\mathbf{P}^m$ in each patch of the pixel-wise temporal difference feature representations. However, $\mathbf{P}^p$ only encodes the more representative information in each patch, resulting in losing the contextual information in bi-temporal images. To this end, the pyramid adaptive average pooling used in~\cite{wu2022p2t} is introduced to capture the global semantic contexts in diverse pooling size and construct a series of pyramid features $\mathbf{P}^a_1, \cdots, \mathbf{P}^a_n$. Such process can be denoted as,
	\begin{equation}
		\begin{split}
			&\mathbf{P}^m = \mathsf{MP}(\mathbf{P}_s), \\
			&\mathbf{P}^a_1 = \mathsf{AP}_1(\mathbf{P}_s), \\
			& \ \ \ ..., \\
			&\mathbf{P}^a_n = \mathsf{AP}_n(\mathbf{P}_s), \\
		\end{split}
	\end{equation}
	where $\mathsf{AP}(\cdot)$ and $\mathsf{MP}(\cdot)$ represent the adaptive average pooling and adaptive max pooling operations, respectively.
	
	Then, the global and local contexts are passed through the corresponding convolution layer for feature transfer and concatenated with memory prototypes, like,
	\begin{equation}
		\begin{split}
			\hat{\mathbf{M}}_s = &\mathsf{Cat}(\mathbf{M}_s, \mathsf{Conv}_1(\mathbf{P}^m), \\
			&\mathsf{Conv}_1(\mathbf{P}^a_1), \cdots \mathsf{Conv}_1(\mathbf{P}^a_n)),
		\end{split}
	\end{equation}
	where $\mathsf{Conv}_1(\cdot)$ denotes a $1\times 1$ convolution layer. $\mathsf{Cat}(\cdot)$ is the feature concatenation operation. $\hat{\mathbf{M}}_s$ is the pixel-augmented memory prototypes. 
	
	With the above augmentation manner, we utilize the cross-attention to aggregate information from the pixel-augmented memory prototypes $\hat{\mathbf{M}_s}$ into $\mathbf{M}_s$. The forward pass of the attention is formulated as,
	\begin{equation}
		\begin{split}
			&\mathbf{M}_{s+1} = \mathsf{softmax}(\frac{\mathbf{M}_{s}\mathbf{W}^{q}(\hat{\mathbf{M}_s}\mathbf{W}^{k})^{\top}}{\hat{\mathbf{M}_s}\mathbf{W}^{v}}), \\
			&\mathbf{M}_{s+1} = {\mathbf{M}_{s}} + {\mathbf{M}_{s+1}}\mathbf{W}^{o},
		\end{split}
	\end{equation}
	where $\mathbf{W}^{q}, \mathbf{W}^{k}, \mathbf{W}^{v}, \mathbf{W}^{o}$ denote four learnable linear projections. The cross-attention updates the memory prototypes with the pixel-level temporal difference representations and models the relationships of diverse memory prototypes, which makes it possible to learn the prototypes of various changes deceptively for different bi-temporal images.
	
	\subsubsection{Memory-to-Pixel Attention}
	After updating the memory prototypes, we apply the memory-to-pixel (M2P) attention, implemented by the cross-attention to query additional information from the memory prototypes to generate more discriminative temporal difference feature representations. The cross attention can be formulated as,
	\begin{equation}
		\begin{split}
			&\mathbf{P}_{s+1} = \mathsf{softmax}(\frac{\mathbf{P}_{s}\mathbf{W}^{q}(\mathbf{M}_{s+1}\mathbf{W}^{k})^{\top}}{\mathbf{M}_{s+1}\mathbf{W}^{v}}), \\
			&\mathbf{P}_{s+1} = {\mathbf{P}_{s}} + {\mathbf{P}_{s+1}}\mathbf{W}^{o},
		\end{split}
	\end{equation}
	
	The cross-attention updates pixel-level temporal difference feature representations with the memory prototypes and makes it aware of the presence of the whole dataset, for better adjusting the diversities of changed and unchanged regions in the embedding space.
	
	\subsection{Patch-level Supervision Scheme}\label{sub:pss}
	After obtaining the updated temporal difference features from the bi-directional attention blocks, we generate the change maps $\mathbf{G}$ with a convolution layer. Then, patch-level annotations are utilized to guide the learning process of the network. To achieve this goal, we introduce a patch-level supervision scheme.
	
	The patch-level supervisions indicate whether a pair of bi-temporal image patches contain changes or not, as defined in Eq.~\eqref{eq:patch_level_annotations}. In paired bi-temporal image patches, we select the most significant pixel to formulate a local-scale change map, which preserves the change information of patch-level label, with an adaptive max pooling operation. 
	\begin{equation}
		\mathbf{Y}^l = \mathsf{MP}(\mathbf{Y}),  \ \mathbf{G}^l = \mathsf{MP}(\mathbf{G}),
	\end{equation}
	where $\mathbf{Y}^l$ and $\mathbf{G}^l$ are the ground-truth local-scale change maps and predicted ones, respectively. Then, the binary cross-entropy (BCE) loss $\mathcal{L}_{bce}$ is exploited to supervise the predicted local-scale change maps as well as classify bi-temporal image patches into changed or unchanged classes, like,
	\begin{equation}
		\begin{split}
			\mathcal{L}_{pcl} &= \mathcal{L}_{bce}(\mathbf{G}^l, \mathbf{Y}^l) \\
			&= \mathbf{G}^l \cdot \mathrm{log}(\mathbf{Y}^l) + (\mathbf{1} - \mathbf{G}^l) \cdot \mathrm{log}(\mathbf{1} -\mathbf{Y}^l)
		\end{split} 
	\end{equation}
	where $\mathcal{L}_{pcl}$ is the patch classification loss. 
	
	In addition to the patch classification loss, we construct an unchanged patch consistency loss to filter the miss-predictions that occur in unchanged patches as,
	\begin{equation}
		\mathcal{L}_{upcl} = \Vert (\mathbf{1} - \mathbf{Y})\cdot(\mathbf{G} - \mathbf{Y})\Vert_1
	\end{equation} 
	where $\Vert \cdot \Vert_1$ is the $\ell_1$ loss. 
	
	Thus, the loss of the patch-level supervision scheme can be formulated as,
	\begin{equation}
		\mathcal{L}_{pss} = \mathcal{L}_{pcl}(\mathbf{G}^l, \mathbf{Y}^l) +  \mathcal{L}_{upcl}(\mathbf{G}, \mathbf{Y})
	\end{equation}
	
	\subsection{Loss Function}
	In our BAB, we extract the most representative feature $\mathbf{P}^m$ from each patch in the paired bi-temporal images. Here, we additionally use the BCE loss to ensure semantic properties of $\mathbf{P}^m$. Suppose the $\mathbf{Q}_s$ is the predicated semantic map of $\mathbf{P}^m$ in $s$-th block of BAB by a single convolution layer, the loss can be represented as,
	\begin{equation}
		\mathcal{L}_{sp} = \mathcal{L}_{bce}(\mathbf{Q}^s, \mathbf{Y}^l)
	\end{equation}
	
	With the loss of the patch-level supervision scheme, the total training loss of MS-Former can be denoted as,
	\begin{equation}
		\mathcal{L} = \sum_{s=1}^S\mathcal{L}_{sp}(\mathbf{Q}^s, \mathbf{Y}^l) + \mathcal{L}_{pss}
	\end{equation}
	where $S$ is the number of BAB in the MS-Former.
	
	\section{Experiments}\label{sec:sec4}
	
	\subsection{Datasets}
	
	\begin{table*}[!htbp]
		\centering
		\caption{Ablation Studies of the proposed method with diverse settings in terms of $\kappa$, IoU, and F1 on BCDD dataset.}
		\label{tab:ablation_studies}
		\resizebox{\textwidth}{!}{%
			\begin{tabular}{@{}cccccccccccccc@{}}
				\toprule[1.5pt]
				\multicolumn{1}{c|}{\multirow{2}{*}{No.}} & \multicolumn{1}{c|}{\multirow{2}{*}{Variants}} & \multicolumn{3}{c|}{$256\times 256$}           & \multicolumn{3}{c|}{$128\times 128$}           & \multicolumn{3}{c|}{$64\times 64$}             & \multicolumn{3}{c}{$32\times 32$} \\ \cmidrule(l){3-14} 
				\multicolumn{1}{c|}{}                     & \multicolumn{1}{c|}{}                          & $\kappa$ & IoU    & \multicolumn{1}{c|}{F1}     & $\kappa$ & IoU    & \multicolumn{1}{c|}{F1}     & $\kappa$ & IoU    & \multicolumn{1}{c|}{F1}     & $\kappa$    & IoU       & F1       \\ \midrule
				\multicolumn{1}{c|}{\#01}                 & \multicolumn{1}{c|}{WCDNet}                    & 0.7332  & 0.5912 & \multicolumn{1}{c|}{0.7431} & 0.8342  & 0.7262 & \multicolumn{1}{c|}{0.8414} & 0.8857  & 0.8028 & \multicolumn{1}{c|}{0.8906} & 0.8980     & 0.8221    & 0.9024   \\ \midrule
				\multicolumn{14}{l}{(a) Bi-directional Attention Block (BAB)}                                                                                                                                                                                                                     \\ \midrule
				\multicolumn{1}{c|}{\#02}                 & \multicolumn{1}{c|}{w/o BAB}                   & 0.7172  & 0.5711 & \multicolumn{1}{c|}{0.7270} & 0.7882  & 0.6621 & \multicolumn{1}{c|}{0.7967} & 0.8640  & 0.7694 & \multicolumn{1}{c|}{0.8697} & 0.8294     & 0.7186    & 0.8363   \\
				\multicolumn{1}{c|}{\#03}                 & \multicolumn{1}{c|}{$N_m=64$}                    & 0.7213  & 0.5768 & \multicolumn{1}{c|}{0.7316} & 0.8360  & 0.7291 & \multicolumn{1}{c|}{0.8433} & 0.8808  & 0.7951 & \multicolumn{1}{c|}{0.8859} & 0.8800     & 0.7939    & 0.8851   \\
				\multicolumn{1}{c|}{\#04}                 & \multicolumn{1}{c|}{$N_m=192$}                   & 0.7164  & 0.5712 & \multicolumn{1}{c|}{0.7271} & 0.8288  & 0.7186 & \multicolumn{1}{c|}{0.8362} & 0.8770  & 0.7893 & \multicolumn{1}{c|}{0.8822} & 0.8814     & 0.7960    & 0.8864   \\
				\multicolumn{1}{c|}{\#05}                 & \multicolumn{1}{c|}{BAB w/o P2M}                   & 0.6612  & 0.5056 & \multicolumn{1}{c|}{0.6716} & 0.8155  & 0.7001 & \multicolumn{1}{c|}{0.8236} & 0.8651  & 0.7712 & \multicolumn{1}{c|}{0.8708} & 0.8230     & 0.7093    & 0.8299   \\
				\multicolumn{1}{c|}{\#06}                 & \multicolumn{1}{c|}{BAB w/o MP}                    & 0.7389  & 0.6002 & \multicolumn{1}{c|}{0.7501} & 0.8150  & 0.6993 & \multicolumn{1}{c|}{0.8230} & 0.8720  & 0.7816 & \multicolumn{1}{c|}{0.8774} & 0.8442     & 0.7399    & 0.8505   \\
				\multicolumn{1}{c|}{\#07}                 & \multicolumn{1}{c|}{BAB w/o AP}                    & 0.7011  & 0.5529 & \multicolumn{1}{c|}{0.7121} & 0.8321  & 0.7232 & \multicolumn{1}{c|}{0.8394} & 0.8748  & 0.7859 & \multicolumn{1}{c|}{0.8801} & 0.8712     & 0.7803    & 0.8766   \\ \midrule
				\multicolumn{14}{l}{(b) Patch-level Supervision Scheme (PSS)}                                                                                                                                                                                                                     \\ \midrule
				\multicolumn{1}{c|}{\#08}                 & \multicolumn{1}{c|}{PSS w/o L1}                    & 0.6863  & 0.5352 & \multicolumn{1}{c|}{0.6973} & 0.8355  & 0.7280 & \multicolumn{1}{c|}{0.8426} & 0.8741  & 0.7849 & \multicolumn{1}{c|}{0.8795} & 0.8872     & 0.8051    & 0.8920   \\
				\multicolumn{1}{c|}{\#09}                 & \multicolumn{1}{c|}{PSS w/o BCE}                   & 0.7235  & 0.5797 & \multicolumn{1}{c|}{0.7339} & 0.8300  & 0.7198 & \multicolumn{1}{c|}{0.8085} & 0.7994  & 0.6763 & \multicolumn{1}{c|}{0.8069} & 0.0063     & 0.0033    & 0.0066   \\
				\multicolumn{1}{c|}{\#10}                 & \multicolumn{1}{c|}{Directly sup}              & 0.2976  & 0.2101 & \multicolumn{1}{c|}{0.3472} & 0.4207  & 0.2977 & \multicolumn{1}{c|}{0.4588} & 0.5646  & 0.4191 & \multicolumn{1}{c|}{0.5906} & 0.6986     & 0.5564    & 0.7150   \\ \bottomrule[1.5pt]
			\end{tabular}%
		}
	\end{table*}
	
	\begin{table*}[!htbp]
		\centering
		\caption{Quantitative comparisons of the proposed method with full supervised change detection methods in terms of $\kappa$, IoU, F1, OA, Rec, and Pre on the LEVIR and SYSU datasets.}
		\label{tab:fully_supervised_results}
		\resizebox{\textwidth}{!}{%
			\begin{tabular}{@{}cc|c|c|c|c|c|c|c|c|c|c@{}}
				\toprule[1.5pt]
				\multicolumn{2}{c|}{}                                   & \multicolumn{6}{c}{Full-supervised methods}                                &                               & \multicolumn{3}{c}{Our weakly-supervised method}    \\ \cmidrule(l){3-12} 
				\multicolumn{2}{c|}{\multirow{-2}{*}{Methods}}         & STANet & L-Unet & SNUNet & DSIFN  & BIT    & TFI-GR                        & A2Net                         & $128\times 128$ & $64\times 64$   & $32\times 32$   \\ \midrule
				\multicolumn{2}{c|}{FLOPs (G)}                          & 25.69  & 34.63  & 246.22 & 164.56 & 16.88  & 19.44                         & 6.02                          & 22.14           & 22.14           & 22.14           \\ \midrule
				\multicolumn{2}{c|}{Params (M)}                         & 12.28  & 8.45   & 27.07  & 50.71  & 3.01   & 28.37                         & 3.78                          & 15.20           & 15.20           & 15.20           \\ \midrule
				\multicolumn{1}{c|}{}                        & $\kappa$ & 0.8839 & 0.8935 & 0.8972 & 0.9090 & 0.9007 & 0.9045                        & {\color[HTML]{00B0F0} 0.9129} & 0.6982 ($\downarrow$0.2147) & 0.7561 ($\downarrow$0.1568) & 0.8034 ($\downarrow$0.1095) \\
				\multicolumn{1}{c|}{}                        & IoU      & 0.8014 & 0.8164 & 0.8220 & 0.8408 & 0.8276 & 0.8336                        & {\color[HTML]{00B0F0} 0.8472} & 0.5571 ($\downarrow$0.2901) & 0.6258 ($\downarrow$0.2214) & 0.6864 ($\downarrow$0.1608) \\
				\multicolumn{1}{c|}{}                        & F1       & 0.8898 & 0.8989 & 0.9023 & 0.9135 & 0.9057 & 0.9093                        & {\color[HTML]{00B0F0} 0.9173} & 0.7156 ($\downarrow$0.2017) & 0.7698 ($\downarrow$0.1475) & 0.8141 ($\downarrow$0.1032) \\
				\multicolumn{1}{c|}{}                        & Rec      & 0.8761 & 0.8915 & 0.8897 & 0.8944 & 0.8940 & 0.8973                        & {\color[HTML]{00B0F0} 0.9059} & 0.8223 ($\downarrow$0.0836) & 0.8606 ($\downarrow$0.0453) & 0.8692 ($\downarrow$0.0367) \\
				\multicolumn{1}{c|}{\multirow{-5}{*}{LEVIR}} & Pre      & 0.9038 & 0.9064 & 0.9154 & 0.9335 & 0.9177 & 0.9215                        & {\color[HTML]{00B0F0} 0.9290} & 0.6334 ($\downarrow$0.2956) & 0.6964 ($\downarrow$0.2326) & 0.7655 ($\downarrow$0.1635) \\ \midrule
				\multicolumn{1}{c|}{}                        & $\kappa$ & 0.7100 & 0.7398 & 0.7391 & 0.7205 & 0.7333 & 0.7902                        & {\color[HTML]{00B0F0} 0.7916} & 0.5979 ($\downarrow$0.1937) & 0.6681 ($\downarrow$0.1235) & 0.7364 ($\downarrow$0.0552) \\
				\multicolumn{1}{c|}{}                        & IoU      & 0.6332 & 0.6662 & 0.6673 & 0.6442 & 0.6584 & {\color[HTML]{00B0F0} 0.7240} & 0.7237                        & 0.5065 ($\downarrow$0.2175) & 0.5794 ($\downarrow$0.1446) & 0.6559 ($\downarrow$0.0681) \\
				\multicolumn{1}{c|}{}                        & F1       & 0.7754 & 0.7996 & 0.8004 & 0.7836 & 0.7940 & {\color[HTML]{00B0F0} 0.8399} & 0.8397                        & 0.6724 ($\downarrow$0.1675) & 0.7337 ($\downarrow$0.1062) & 0.7922 ($\downarrow$0.0477) \\
				\multicolumn{1}{c|}{}                        & Rec      & 0.7430 & 0.7808 & 0.7979 & 0.7511 & 0.7668 & {\color[HTML]{00B0F0} 0.8437} & 0.8224                        & 0.5538 ($\downarrow$0.2899) & 0.6314 ($\downarrow$0.2123) & 0.7181 ($\downarrow$0.1256) \\
				\multicolumn{1}{c|}{\multirow{-5}{*}{SYSU}}  & Pre      & 0.8108 & 0.8195 & 0.8030 & 0.8190 & 0.8232 & 0.8361                        & {\color[HTML]{00B0F0} 0.8577} & 0.8556 ($\downarrow$0.0021) & 0.8755 ($\downarrow$0.0178) & 0.8832 ($\uparrow$0.0255) \\ \bottomrule[1.5pt]
			\end{tabular}%
		}
	\end{table*}
	
	We perform experiments on three widely used CD datasets, including BCDD~\footnote{http://study.rsgis.whu.edu.cn/pages/download/building\_dataset.html}, LEVIR~\cite{chen2020spatial}, and SYSU~\cite{shi2021deeply}. The detailed information is given as follows, 
	
	\textbf{BCDD}: It is a high resolution (0.3m) urban building CD dataset, consisting of a pair of aerial images that size is $32507\times 15354$, collected in 2012 and 2016. We leverage the data processed in BGMix~\cite{huang2023background}, which crops the bi-temporal images into $256\times 256$ patches. Then, $90\%$ and $10\%$ samples of the dataset are randomly selected for training and testing. Finally, the cropped dataset consists of 1890 changed and 5544 unchanged paired image patches.
	
	\textbf{LEVIR}: It is a high resolution (0.3m) building CD dataset, which exists of 637 pairs of bi-temporal remote sensing images of $1024\times 1024$ spatial size. The method in~\cite{chen2020spatial} officially split the dataset into 7:1:2 for training, validation, and testing. We crop the original images into $256\times 256$ none-overlapping patches, and obtained a total of 7120/1024/2048 image pairs for training/validation/testing, respectively.
	
	\textbf{SYSU}: This dataset consists of 20000 pairs of bi-temporal remote sensing image patches with $256\times 256$ spatial size and $0.5$m spatial resolution. The ratios of training, validation, and testing are officially set to $6:2:2$. This dataset contains various types of complex change scenes, including road expansion, newly built urban buildings, change of vegetation, suburban dilation, and groundwork before construction.
	
	\subsection{Evaluation Metrics}
	In our experiments, five widely used metrics, namely Kappa coefficient ($\kappa$), intersection over union (IoU), F1-score (F1), recall (Rec), and precision (Pre) are employed to evaluate the performance of the CD task. The detailed calculations of five metrics can be found in ~\cite{Li2022cd, li2023lightweight}.
	
	\subsection{Implementation details}
	In this paper, the Resnet18~\cite{he2016deep} pre-trained on ImageNet is exploited as the image encoder to extract bi-temporal features from bi-temporal images, for a fair comparison. Then, the decoder is set as the UperNet~\cite{xiao2018unified}, which is widely used for aggregate multi-level features in the semantic segmentation task. The memory length $N_m$ and the number of BAB are set as 128 and 3, respectively. As done in~\cite{wu2022p2t}, the pyramid adaptive average pooling ratios are set as 12, 16, 20, and 24.
	
	The proposed method is implemented via the Pytorch toolbox~\cite{paszke2019pytorch} and performed on a single Nvidia Titan V GPU. The Adam optimizer~\cite{2014Adam}, in which the momentum, weight decay, parameters $\beta_1$ and $\beta_2$ are respectively set to 0.9, 0.0001, 0.9, and 0.99, to optimize the network. Then, the $poly$ learning scheme is applied to adjust the learning rate as $(1 - \frac{cur\_iteration}{max\_iteration} )^{power} \times lr$, where $power$ and $max\_iteration$ are set as 0.9 and 40000, respectively. We set the initial learning rate as 0.0005 and the batch size as 32. Random flipping and temporal exchanging are employed on the image patches for data augmentation.
	
	\subsection{Ablation Studies}
	In this part, we perform ablation studies on the BCDD dataset to explore the effectiveness of diverse components in the proposed method. 
	
	\subsubsection{Effectiveness of bi-directional attention block (BAB)}
	In the proposed method, memory prototypes are introduced to preserve the attributes of changed and unchanged regions. The Bi-directional Attention Block (BAB) is proposed to learn the memory prototypes from pixel-wise temporal difference feature representations and enhance the temporal difference representation with the updated memory prototypes. To validate the effectiveness of BAB, we directly remove the BAB from the network and term the method as w/o BAB (\#02 in TABLE~\ref{tab:ablation_studies}). From the results in TABLE~\ref{tab:ablation_studies}, the performance of w/o BAB decreases about 1.61, 4.47, 2.09, and 6.61 percentages in terms of F1 compared with \#01 on the BCDD dataset in $256\times 256$, $128\times 128$, $64\times 64$, and $32\times 32$ patch size settings, respectively. This indicates the effectiveness of the proposed BAB and emphasizes the importance of memory-augmented representation learning manner in boosting performance for the weakly-supervised CD task.
	
	The length of memory prototypes $N_m$ is a parameter in the BAB. We search the parameter in the range of 64, 128, 192. From the results in TABLE \ref{tab:ablation_studies}, we observe that \#01 outperforms \#03 and \#04 on the BCDD dataset. Thus, the length of memory prototypes is set to 128 in the proposed method.
	
	To further study the effectiveness of diverse components in BAB, we construct three comparators, i.e., BAB w/o P2M, BAB w/o MP, and BAB w/o AP (\#05, \#06, \#07 in TABLE~\ref{tab:ablation_studies}). For BAB w/o P2M, we drop out the pixel-to-memory attention and just utilize the self-attention to self-update the memory prototypes. As shown in TABLE~\ref{tab:ablation_studies}, the performance of BAB w/o P2M declines about 7.15, 1.78, 1.89, and 7.29 percentages measured by F1 compared with \#01 on the BCDD dataset under four patch size settings, respectively.
	Thus, leveraging the temporal difference information to update the memory prototypes is important in the proposed method. As to BAB w/o MP and BAB w/o AP, we respectively remove the global and local representation feature extractors in the pixel to memory attention. From the results in TABLE~\ref{tab:ablation_studies}, such two methods both decrease a lot compared with \#01 on the BCDD dataset. Therefore, both global and local representation feature extractors play a key role in updating the memory prototypes and consequently boost the change detection performance.
	
	\subsubsection{Effectiveness of patch-level supervision scheme (PSS)}
	The Patch-level Supervision Scheme (PSS) is formulated to guide the network learning knowledge from the patch-level annotations.
	In PSS, the patch classification loss and consistency loss are jointly utilized. To verify their effectiveness, we respectively remove the patch classification loss and consistency loss termed as PSS w/o pcl and PSS w/o cl (\#08 and \#09 in TABLE \ref{tab:ablation_studies}). As seen from the results, the change detection performance of PSS w/o pcl and PSS w/o cl methods both decrease significantly on BCDD datasets. Thus, we can observe that the patch classification loss and consistency loss can improve the change detection performance, indicating the effectiveness of PSS. In addition, we formulate a method that directly uses the patch-level annotation to supervise the network via binary cross entropy loss as w DS (\#10 in TABLE \ref{tab:ablation_studies}). Compared with PSS, w DS method reduces the performance a lot.  
	
	\subsection{Comparison With Weakly-supervised CD Methods}
	We compare the proposed model with ten state-of-the-art remote sensing change detection approaches, including WCDNet~\cite{andermatt2020weakly}, WS-C~\cite{kalita2021land}, CAM~\cite{wu2023fully}, FCD-GAN~\cite{wu2023fully}, 
	BGMix~\cite{huang2023background}, TransWCD~\cite{zhao2023exploring}, and TransWCD-DL~\cite{zhao2023exploring}. The change detection performance of various methods in terms of $\kappa$, IoU, F1, Rec, and Pre on the BCDD dataset is given in TABLE~\ref{tab:weakly_supervised_CD}. 
	
	\begin{table}[!htbp]
		\centering
		\caption{Quantitative comparisons of the proposed method with other image-level supervised change detection methods in terms of $\kappa$, IoU, F1, OA, Rec, and Pre on the BCDD dataset.}
		\label{tab:weakly_supervised_CD}
		\resizebox{0.48\textwidth}{!}{%
			\begin{tabular}{@{}c|c|c|c|c|c|c@{}}
				\toprule[1.5pt]
				Methods               & Patch-size      & $\kappa$ & IoU    & F1     & Rec    & Pre    \\ \midrule
				WCDNet                & $256\times 256$ & -        & 0.2210 & 0.3930 & -      & -      \\
				WS-C                  & $256\times 256$ & -        & 0.1937 & 0.3245 & 0.2387 & 0.5067 \\
				CAM                   & $256\times 256$ & -        & 0.3755 & 0.5460 & 0.4940 & 0.6102 \\
				FCD-GAN               & $256\times 256$ & -        & 0.3932 & 0.5645 & 0.4893 & 0.6678 \\
				BGMix                 & $256\times 256$ & -        & 0.4270 & 0.6240 & -      & -      \\
				TransWCD              & $256\times 256$ & -        & 0.5236 & 0.6873 & 0.7534 & 0.6319 \\
				TransWCD-DL           & $256\times 256$ & -        & 0.5619 & 0.7195 & 0.6446 & 0.8142 \\ \midrule
				\multirow{4}{*}{Ours} & $256\times 256$ & 0.7332   & 0.5912 & 0.7431 & 0.6501 & 0.8672 \\
				& $128\times 128$ & 0.8342   & 0.7262 & 0.8414 & 0.8483 & 0.8346 \\
				& $64\times 64$   & 0.8857   & 0.8028 & 0.8906 & 0.8854 & 0.8958 \\
				& $32\times 32$   & 0.8980   & 0.8221 & 0.9024 & 0.8983 & 0.9065 \\ \bottomrule[1.5pt]
			\end{tabular}%
		}
	\end{table}
	
	As shown in TABEL~\ref{tab:weakly_supervised_CD}, our proposed method exhibits superior performance compared with the other seven image-level annotations supervised approaches. For instance, our proposed method obtains about 2.93, 2.36, 0.55, and 5.30 percentages higher performances in terms of IoU, F1, Rec, and Pre than the second-best approach (TransWCD-DL) under $256\times 256$ patch size setting, respectively. Meanwhile, our proposed method consistently achieves superior change detection results than other approaches when the patch size is set to $128\times 128$, $64\times 64$, and $32\times 32$. The above observations strongly demonstrate the effectiveness and superiority of the proposed method. In addition, when the patch size decreases, our proposed method can capture more accurate change detection results, verifying the effectiveness of the idea of employing patch-level annotations to achieve the trade-offs between performance and costs.
	
	\subsection{Comparison With Fully-supervised CD Methods}
	
	We compare the proposed model with several state-of-the-art remote sensing change detection approaches, including
	STANet~\cite{chen2020spatial}, L-Unet~\cite{papadomanolaki2021deep}, SNUNet~\cite{fang2021snunet}, DSIFN~\cite{zhang2020deeply}, BIT~\cite{chen2021remote}, TFI-GR~\cite{Li2022cd}, and A2Net~\cite{li2023lightweight}.  
	
	\subsubsection{Quantitative comparisons}
	The quantitative evaluation results in terms of $\kappa$, IoU, F1, Rec, and Pre of our weakly supervised MS-Former and other fully supervised change detection methods are reported in TABLE~\ref{tab:fully_supervised_results}. The model parameters (Params), and computation costs (FLOPs) related to all approaches are also given in TABLE~\ref{tab:fully_supervised_results}. From the results, we observe that there are about 20.17 and 16.75 percentage gaps between our MS-Former with $128\times 128$ patch-size patch-level annotations and the best fully supervised change detection performers measured by F1 on LEVIR and SYSU datasets. Our MS-Former with $32\times 32$ patch-size patch-level annotations can significantly narrow the gap to 10.32 and 4.77 percentages in terms of F1 on LEVIR and SYSU datasets. It should be noted that the change detection performance gap keeps getting smaller as the patch size goes smaller. Such observations strongly indicate the effectiveness of the patch-level annotation and our MS-Former method.
	
	\subsubsection{Visual comparisons}
	
	\begin{figure*}[!htbp]
		\centering
		\includegraphics[width=0.9\textwidth]{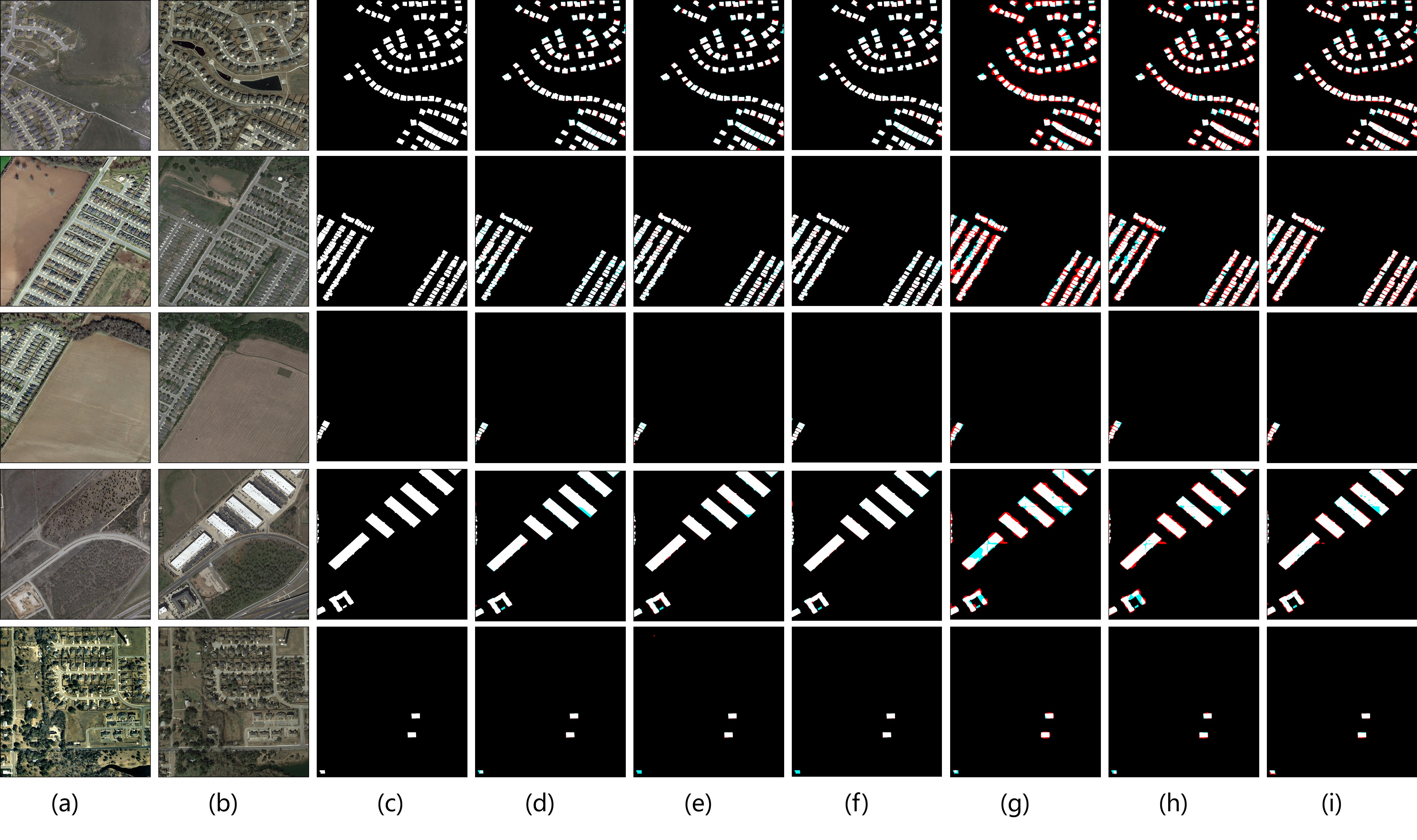}
		\caption{Visual comparisons of the proposed MS-Former and other methods on the LEVIR dataset. (a) $t_1$ images; (b) $t_2$ images; (c) Ground-truth; (d) BIT; (e) TFI-GR; (f) A2Net; (g) our MS-Former with $128 \times 128$ patch size; (h) our MS-Former with $64 \times 64$ patch size; (i) our MS-Former with $32 \times 32$ patch size. The rendered colors represent true positives (white), false positives ({\color[HTML]{FF0000} red}), true negatives (black),  and false negatives ({\color[HTML]{00B0F0} blue}).}
		\label{fig:vis_cd_LEVIR}
	\end{figure*}
	
	\begin{figure*}[!htbp]
		\centering
		\includegraphics[width=0.9\textwidth]{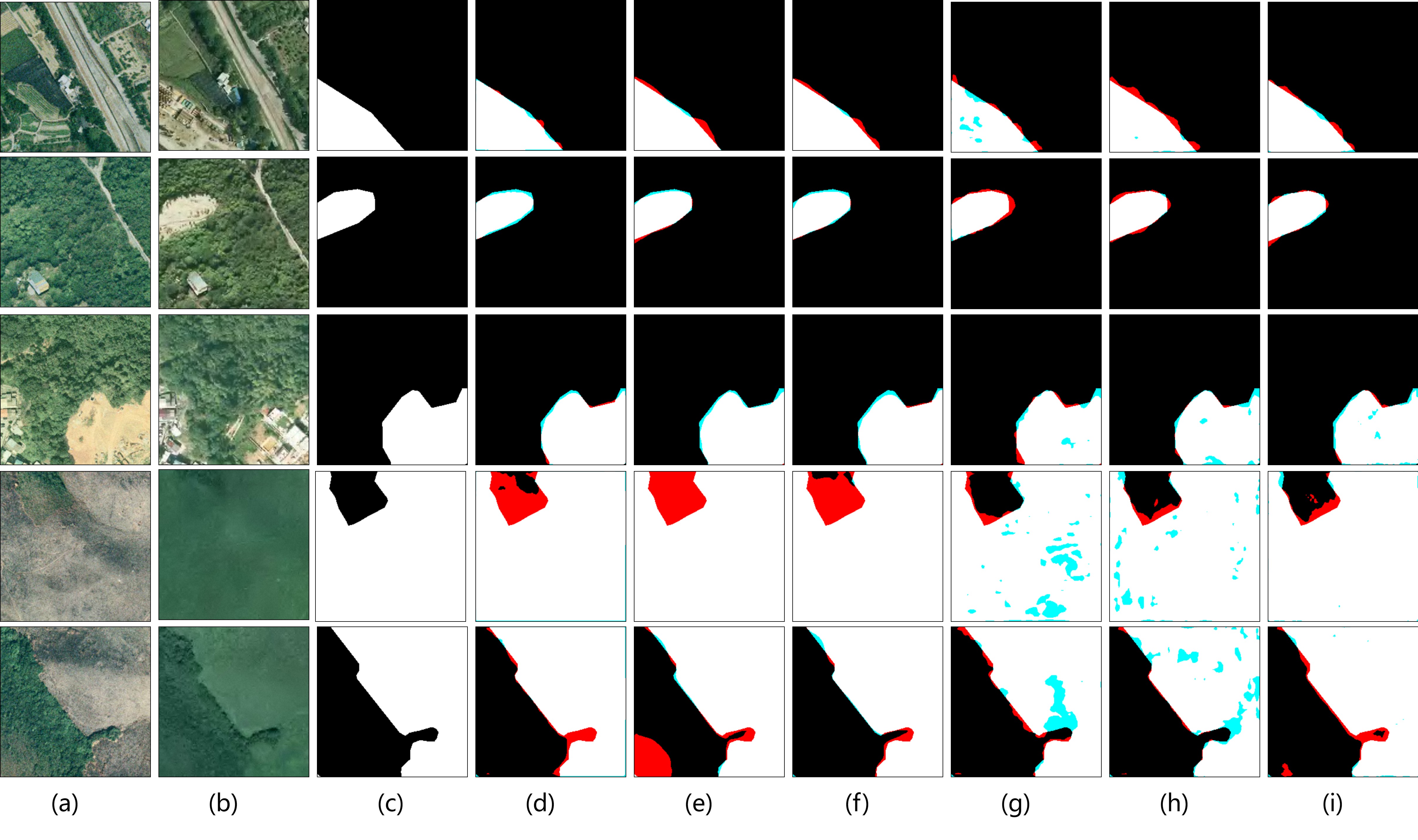}
		\caption{Visual comparisons of the proposed MS-Former and other methods on the SYSU dataset. (a) $t_1$ images; (b) $t_2$ images; (c) Ground-truth; (d) BIT; (e) TFI-GR; (f) A2Net; (g) our MS-Former with $128 \times 128$ patch size; (h) our MS-Former with $64 \times 64$ patch size; (i) our MS-Former with $32 \times 32$ patch size. The rendered colors represent true positives (white), false positives ({\color[HTML]{FF0000} red}), true negatives (black),  and false negatives ({\color[HTML]{00B0F0} blue}).}
		\label{fig:vis_cd_SYSU}
	\end{figure*}
	
	To intuitively compare our MS-Former and other methods, we present the visual comparison results of BIT, TFI-GR, A2Net, and our MS-Former on LEVIR+ and BCDD datasets in Figs.~\ref{fig:vis_cd_LEVIR} and~\ref{fig:vis_cd_SYSU}, respectively. White, red, black, and blue are used to indicate true positives, false positives, true negatives, and false negatives, respectively, for better visualization. Based on the visual results in Figs. \ref{fig:vis_cd_LEVIR} and \ref{fig:vis_cd_SYSU}, we observe that the proposed method demonstrates superiority in the following aspects:
	
	\begin{table*}[!htbp]
		\centering
		\caption{Quantitative comparisons of diverse approaches in terms of $\kappa$, IoU, F1, OA, Rec, and Pre on the GVLM dataset.}
		\label{tab:weakly_supervised_LS}
		\resizebox{\textwidth}{!}{%
			\begin{tabular}{@{}c|c|c|c|c|c|c|c|c|c|c|c|c@{}}
				\toprule
				& \multicolumn{3}{c|}{Single-temporal full-supervised methods} & \multicolumn{6}{c|}{Bi-temporal full-supervised methods}                                                                  & \multicolumn{3}{c}{Our weakly-supervised method}                                     \\ \cmidrule(l){2-13} 
				\multirow{-2}{*}{Datasets} & SINet-V2            & GeleNet            & BASNet            & DSIFN  & BIT    & MSCANet & TFI-GR                        & A2Net                         & AR-CDNet                      & $128\times 128$            & $64\times 64$              & $32\times 32$              \\ \midrule
				$\kappa$                   & 0.8747              & 0.8774             & 0.8759            & 0.8566 & 0.8851 & 0.8893  & 0.8974                        & 0.8916                        & {\color[HTML]{00B0F0} 0.8987} & 0.7098($\downarrow$0.1889) & 0.8049($\downarrow$0.0938) & 0.8251($\downarrow$0.0736) \\
				IoU                        & 0.7906              & 0.7946             & 0.793             & 0.7638 & 0.8067 & 0.8131  & 0.8256                        & 0.8162                        & {\color[HTML]{00B0F0} 0.8275} & 0.5705($\downarrow$0.2570) & 0.6915($\downarrow$0.1360) & 0.7191($\downarrow$0.1084) \\
				F1                         & 0.8831              & 0.8855             & 0.8846            & 0.8661 & 0.893  & 0.8969  & 0.9045                        & 0.8988                        & {\color[HTML]{00B0F0} 0.9056} & 0.7265($\downarrow$0.1791) & 0.8176($\downarrow$0.0880) & 0.8366($\downarrow$0.0689) \\
				Rec                        & 0.8563              & 0.8571             & 0.8947            & 0.8324 & 0.8936 & 0.9021  & {\color[HTML]{00B0F0} 0.9069} & 0.8749                        & 0.9031                        & 0.6258($\downarrow$0.2811) & 0.7783($\downarrow$0.1286) & 0.8009($\downarrow$0.1060) \\
				Pre                        & 0.9116              & 0.9159             & 0.8746            & 0.9026 & 0.8925 & 0.8918  & 0.9021                        & {\color[HTML]{00B0F0} 0.9241} & 0.9082                        & 0.8659($\downarrow$0.0582) & 0.8611($\downarrow$0.0630) & 0.8755($\downarrow$0.0486) \\ \bottomrule
			\end{tabular}%
		}
	\end{table*}
	\begin{figure*}[!htbp]
		\centering
		\includegraphics[width=0.9\textwidth]{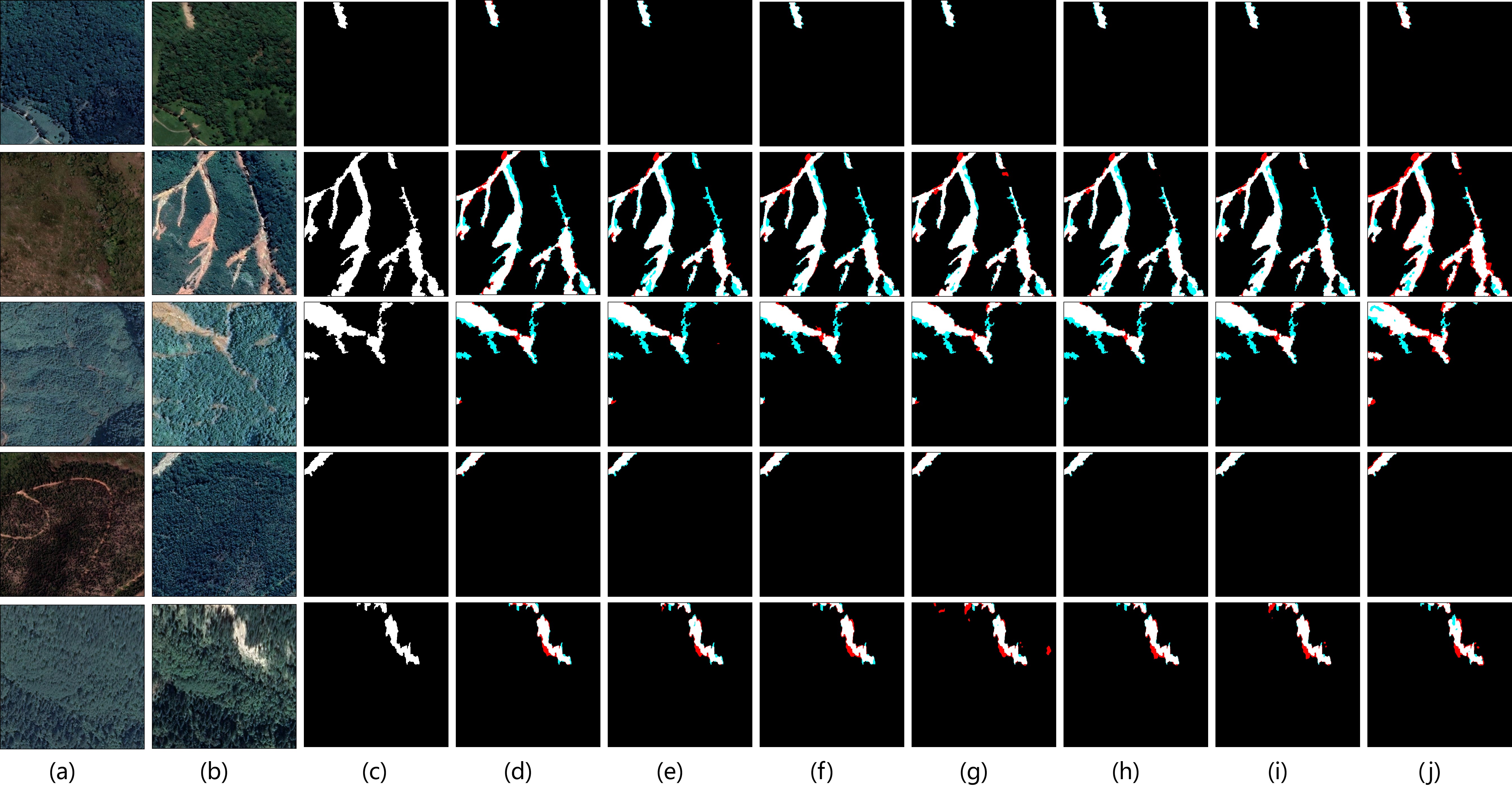}
		\caption{Visual comparisons of the proposed MS-Former and other methods on the GVLM dataset. (a) $t_1$ images; (b) $t_2$ images; (c) Ground-truth; (d) SINet-V2; (e) GeleNet; (f) BASNet; (g) TFI-GR; (h) A2Net; (i) AR-CDNet; (i) our MS-Former with $32 \times 32$ patch size. The rendered colors represent true positives (white), false positives ({\color[HTML]{FF0000} red}), true negatives (black),  and false negatives ({\color[HTML]{00B0F0} blue}).}
		\label{fig:vis_cd_GVLM}
	\end{figure*}
	
	The visual results in Fig.~\ref{fig:vis_cd_LEVIR} demonstrate that the proposed MS-Former generally identifies the changed regions and effectively eliminates the irrelevant pseudo changes caused by background clutters in bi-temporal images. And, the prediction errors mainly lie in the boundary of changed buildings due to lacking fine-gained supervision. In addition, with the patch size decreasing, our MS-Former obtains more accurate change detection results. In Fig.~\ref{fig:vis_cd_SYSU}, it can be observed that BIT, TFI-GR, and A2Net fail to detect the entire changed regions. On the other hand, the proposed MS-Former with $32\times 32$ patch size patch-level annotations can identify the changed objects with fine content integrity. The above comparable performance of our approach is attributed to two main reasons. Firstly, we employ a memory-supported transformer to recover the fine-grained change information from the bi-temporal image patches, and we jointly utilize the global memory prototypes to comprehensively augment the temporal difference features to generate accurate change results. Secondly, our method employs a patch-level supervision scheme to guide the MS-Former learning from the patch-level annotations, resulting in precise change detection results.
	
	\subsection{Downstream Application on Landslide Detection}
	As one of the most widespread disasters, landslides may cause mass human casualties and considerable economic losses every year~\cite{casagli2023landslide}. Thus, detecting such events is an important procedure in landslide hazard monitoring. With the development of earth observation technology, identifying Landslides with the help of very-high-resolution remote sensing imagery has attracted increasing attention in recent years~\cite{ghorbanzadeh2022landslide4sense}. 
	
	In this part, we assess the generalization capabilities of our proposed MS-Former on the landslide detection task with patch-level annotations. To this end, we perform experiments on the carefully collected GLVM~\cite{zhang2023cross} dataset with nine approaches. GLVM is a large-scale and challenging landslide detection dataset, in which the bi-temporal images are collected from 17 diverse regions in the wide world, e.g., Vietnam,  Zimbabwe, New Zealand, and so on. In our experiments, we cropped the bi-temporal images into $512 \times 512$ patches and obtained 1389, 199, and 389 paired images for training, validation, and testing, respectively. The compared methods consist of three single-temporal fully supervised ones (SINet-V2~\cite{fan2021concealed}, GeleNet~\cite{li2023salient}, BASNet~\cite{bo2022basnet}), and six bi-temporal fully supervised ones (DSIFN~\cite{zhang2020deeply}, BIT~\cite{chen2021remote}, MSCANet~\cite{Liu_2022_CD}, TFI-GR~\cite{Li2022cd}, A2Net~\cite{li2023lightweight}, and AR-CDNet~\cite{li2023towards}). The quantitative comparisons of diverse approaches in terms of $\kappa$, IoU, F1, OA, Rec, and Pre on the GVLM dataset are reported in TABLE~\ref{tab:weakly_supervised_LS}. Fig.~\ref{fig:vis_cd_GVLM} shows the visualization results generated by different methods. From the above results, we observe that our proposed MS-Former with patch-level supervision can obtain considerable performance compared with other strong fully supervised methods in the landslide detection task.

	\section{Conclusion}\label{sec:sec5}
	In this work, we introduce a novel memory-supported transformer for weakly supervised change detection with patch-level annotations that is capable of achieving the trade-offs between change detection performance and label annotation costs. The proposed method comprises a bi-directional attention block and a patch-level supervision scheme, which respectively aim to achieve accurate change detection from the feature representation learning and loss function. The experimental results obtained on three high spatial resolution remote sensing change detection datasets demonstrate that the proposed approach outperforms state-of-the-art weakly supervised change detection methods, and obtains considerable performance compared with state-of-the-art fully supervised change detection approaches. We hope that MS-Former will serve as a solid baseline and help ease future research in weakly supervised change detection.
	
	\ifCLASSOPTIONcompsoc
	\section*{Acknowledgments}
	\else
	\section*{Acknowledgment}
	\fi

	The authors wish to gratefully acknowledge the anonymous reviewers for their constructive comments on this paper.

	\ifCLASSOPTIONcaptionsoff
	\newpage
	\fi
	
	\bibliographystyle{IEEEtran}
	\bibliography{IEEEabrv,References}

\end{document}